\documentclass[12pt]{l4dc2022}

\usepackage{graphics} 
\usepackage{mathptmx} 
\usepackage{times} 
\usepackage{amsmath} 
\usepackage{amssymb}  
\usepackage{algorithmic}
\usepackage{algorithm}

\usepackage{enumitem}
\usepackage{balance}
\usepackage{soul,xcolor}
\usepackage{caption}
\usepackage{multirow}

\title[Adaptive Mimic]{Adaptive Mimic: Deep Reinforcement Learning of Parameterized Bipedal Walking from Infeasible References}
\usepackage{times}
\author{%
 \Name{Chong Zhang} \Email{chong-zh18@mails.tsinghua.edu.cn}\\
 \addr Department of Precision Instrument, Tsinghua University, Beijing, 100084 China.
 \AND
 \Name{Qi Wu} \Email{wuqi19@mails.tsinghua.edu.cn}\\
 \addr Department of Mechanical Engineering, Tsinghua University, Beijing, 100084 China.
  \AND
 \Name{Liqian Ma} \Email{mlq19@mails.tsinghua.edu.cn}\\
 \addr Department of Mechanical Engineering, Tsinghua University, Beijing, 100084 China.
  \AND
 \Name{Hongyuan Su} \Email{suhy19@mails.tsinghua.edu.cn}\\
 \addr Department of Electronic Engineering, Tsinghua University, Beijing, 100084 China.%
}

\begin{document}

\maketitle

\begin{abstract}
Not until recently, robust robot locomotion has been achieved by deep reinforcement learning (DRL). However, for efficient learning of parametrized bipedal walking, developed references are usually required, limiting the performance to that of the references. In this paper, we propose to design an adaptive reward function for imitation learning from the references. The agent is encouraged to mimic the references when its performance is low, while to pursue high performance when it reaches the limit of references. We further demonstrate that developed references can be replaced by low-quality references that are generated without laborious tuning and infeasible to deploy by themselves, as long as they can provide a priori knowledge to expedite the learning process.
\end{abstract}

\section{Introduction}
\label{sec:intro}

It has been challenging to design a stable gait for bipedal robots to walk due to their complex dynamics, and much has been explored on modeling and control of bipedal robots. Traditional methods are based on dynamic models and use ZMP (\cite{vukobratovic2004zero}) or Capture Point (\cite{englsberger2011bipedal}) as the criteria for stability, which has led to successful control of many humanoid robots (\cite{huang2001planning}; \cite{hong2013real}; \cite{nelson2012petman}). However, such methods have inherent defects that they typically rely on model-based strategies which are difficult to obtain. To make matters worse, the region of reasonable foot locations is often limited, which makes the robot show little agility. Designers may also need gait libraries to resist violent perturbations and walk in multiple speeds (\cite{xie2018feedback}). In addition, the generalization of these methods can be difficult. When designing new motions other than simple walking, new physical models must be put forward, and complex situations of contact points are hard to deal with. 

Model-free Deep Reinforcement Learning (DRL) provides a way to design controllers for omnidirectional walking without explicit knowledge of complex robot dynamics. DRL also enables the robot to achieve difficult motions like dribbling a ball or following a rough path (\cite{peng2017deeploco}). It has been proved effective in simulation (\cite{lee2019scalable}; \cite{liu2018learning}; \cite{castillo2019reinforcement}; \cite{li2019using}). A single control policy learned from DRL can be applied for locomotion in all directions and different speeds (\cite{rodriguez2021deepwalk}), and can be transferred to real robots after certain sim2real operations (\cite{lee2020learning}; \cite{li2021reinforcement}). 

Typically, there are two kinds of DRL-based approaches, according to the use of references. Approaches without references train the agent directly through trial-and-error, requiring a large amount of training and tuning, and the quality of motions is not guaranteed (\cite{peng2018deepmimic}). An alternative is to track expert motions. While imitation in physics-based model is difficult to implement (\cite{lee2010data}), imitation in DRL has shown much success. 

While researches by (\cite{xie2018feedback}), \cite{peng2018deepmimic}), (\cite{peng2020learning}) and (\cite{li2021reinforcement}) have managed to achieve imitation in DRL, we find that their implementations rely on well-performing references and require motion capture data or laborious tuning of expert controllers. Moreover, in (\cite{li2021reinforcement}), the robot could not walk much faster than the maximum speed of the gait library despite the aggressive commands. Therefore, in this paper, we propose an Adaptive Mimic method that can not only decrease tuning efforts but also take advantage of simple and probably unreliable references.
 
\begin{figure}[t]
  \centering
  \includegraphics[width=150mm]{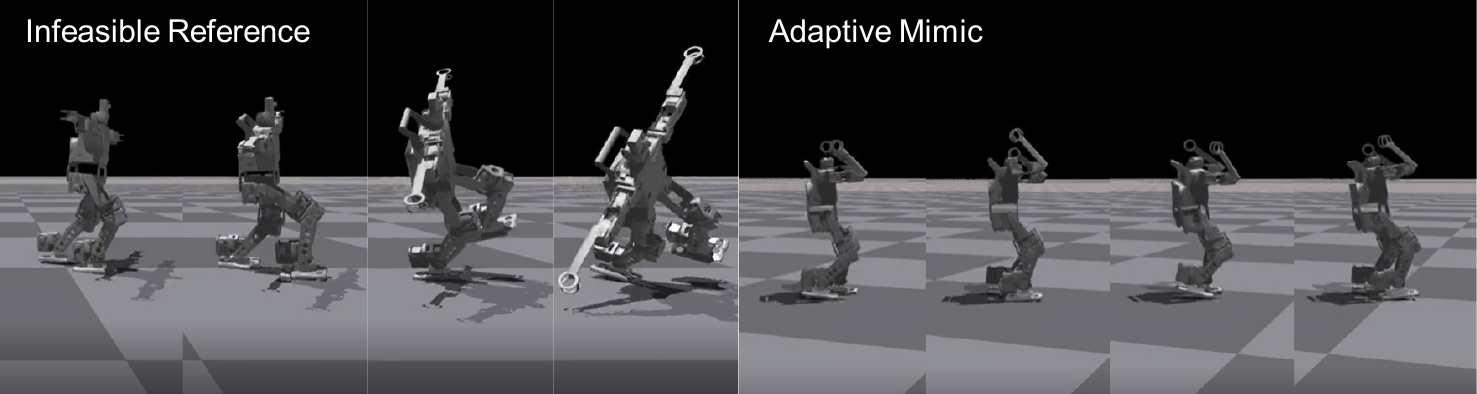}
  \caption{Snapshots of an infeasible reference and the learned gait through our proposed Adaptive Mimic. Although such a reference is unreliable and can not be directly deployed, our method can help the robot learn stable gait from the reference.}
  \label{fig:coverfig}
  \vspace{-1em}
\end{figure}

\subsection{Related Work}
\label{subsec:related work}
Regarding bipedal walking, there are several implementations. One traditional way is the Zero-Moment Point (ZMP). When the biped is dynamically stable, the ZMP should locate inside the support polygon. Therefore, controllers can be designed to generate the desired ZMP using preview control (\cite{kajita2003biped}), convolution sum (\cite{kim2007walking}) and other algorithms (\cite{takanishi1989realization}). Capture Point (CP) is also used in the design of controllers. Capture Point is the point where another step of the robot can stabilize itself. It was originally used in push recovery but was later introduced to gait generation (\cite{englsberger2011bipedal}).  Another approach is to use Hybrid Zero Dynamic (HZD). The main idea of HZD is to use virtual constraints for desired status (\cite{westervelt2018feedback}; \cite{nguyen2020dynamic}). It can be then parameterized by Bezier polynomials and generate a periodic gait. An offline gait library can be generated by HZD and then deployed online (\cite{reher2016realizing}).

Regarding DRL, (\cite{peng2018deepmimic}) proposed DeepMimic, an example-guided DRL framework of complex physics-based skills with which bipeds can do backflips, cartwheels, and rolls. They also proposed a hierarchical DRL approach in (\cite{peng2017deeploco}) to achieve robust gaits and complex locomotion tasks. (\cite{li2021reinforcement}) proposed a framework that combines a gait library from HZD with DRL and implemented it on 3D bipedal locomotion. On the other hand, DRL without references were also brought up. (\cite{rodriguez2021deepwalk}) proposed the DeepWalk framework that does not require references for omnidirectional locomotion. (\cite{rudin2021learning}) proposed a game-inspired curriculum by using massive parallelism on GPU, which effectively shortened the training time. 

\subsection{Motivation and Our Contribution}
In this paper, we aim to find a way to reduce the efforts on tuning for DRL of bipedal walking from references. In our opinion, references should only serve to inspire the agent to achieve a primitive walking pattern, e.g., the robot should lift its foot off the ground and alternate the positions of feet periodically, no matter the velocity or stability achieved. Even infeasible references can be instructive for the agent. Therefore, the process of generating references can be greatly simplified. After learning the primitive walking pattern, the agent should pursue more the performance, but less the imitation. To this end, we propose to automatically adapt the weights of reward terms for performance and imitation during the learning process.

Our contributions in this paper include:
\begin{itemize}
    \item an Adaptive Mimic formulation to mitigate the contradiction between imitation and performance when learning biped locomotion from unreliable references;
    \item showing that laborious manual tuning or high cost for references can be eliminated by imitation learning from a casually designed reference with low quality.
\end{itemize}

With our method, the requirements for the quality of references are greatly reduced, leading to simplification of the reference generation progress. Time consumed on parameter tuning for references and adjustment of reward function is reduced. Better performance and higher efficiency is achieved compared to DeepMimic formulation and direct DRL without reference.

\section{Adaptive Mimic}
\label{sec:adamimic}
Imitation learning of robot behaviors from references, as is formulated in DeepMimic (\cite{peng2018deepmimic}), is based on the reward function. It can be more specifically expressed as
\begin{equation}
    r_t = \omega^I r_t^I + \omega^P r_t^P + \omega^R r_t^R + \omega^O r_t^{O},
    \label{eq:dpm form}
\end{equation}
where $r_t^I$ is the imitation reward, $r_t^P$ is the performance reward that is specified for certain tasks, $r_t^R$ is the regularization reward that encourages realistic movements, $r_t^{O}$ is other reward terms that can facilitate learning, and $\omega^I, \omega^P, \omega^R, \omega^O$ are the weights. Usually, these weights are constants.

However, in the case of low-quality and infeasible references, the imitation reward and the performance reward terms can greatly contradict each other, which can lead to low efficiency and even failures of learning. Yet we expect the robot to learn certain patterns from the references.

To solve this problem, we propose the adaptive mimic (AdaMimic) formulation that can automatically adapt the weights to balance the pursuit of performance and imitation. Specifically, we reformulate Eq. (\ref{eq:dpm form}) as
\begin{equation}
r_t = (1-\omega^P_t) r_t^I + \omega^P_t r_t^P + \omega^R r_t^R + \omega^O r_t^{O},
    \label{eq:ada form}
\end{equation}
with
\begin{equation}
\omega^p_t = \omega^P(r_t^I,r_t^P),
    \label{eq:ada omega}
\end{equation}
i.e., the weight $\omega^P$ is described as a function of $r_t^I$ and $r_t^P$. For convergence of learning process, we further limit the bounds of $r_t^I$ and $r_t^P$:
\begin{equation}
    \label{eq:R star def}
    0\le r_t^I\le R^\star,  \ 0\le r_t^P\le R^\star, 
\end{equation}
where $R^\star$ is a positive real number. We recommend the relationship below:
\begin{equation}
    \label{eq:ideal case}
    \max r_t^I = \max r_t^P = R^\star,
\end{equation}
which is naturally suitable for reward terms in the form of radial basis functions (RBF).

\begin{figure}

  \centering
  \includegraphics[width=120mm]{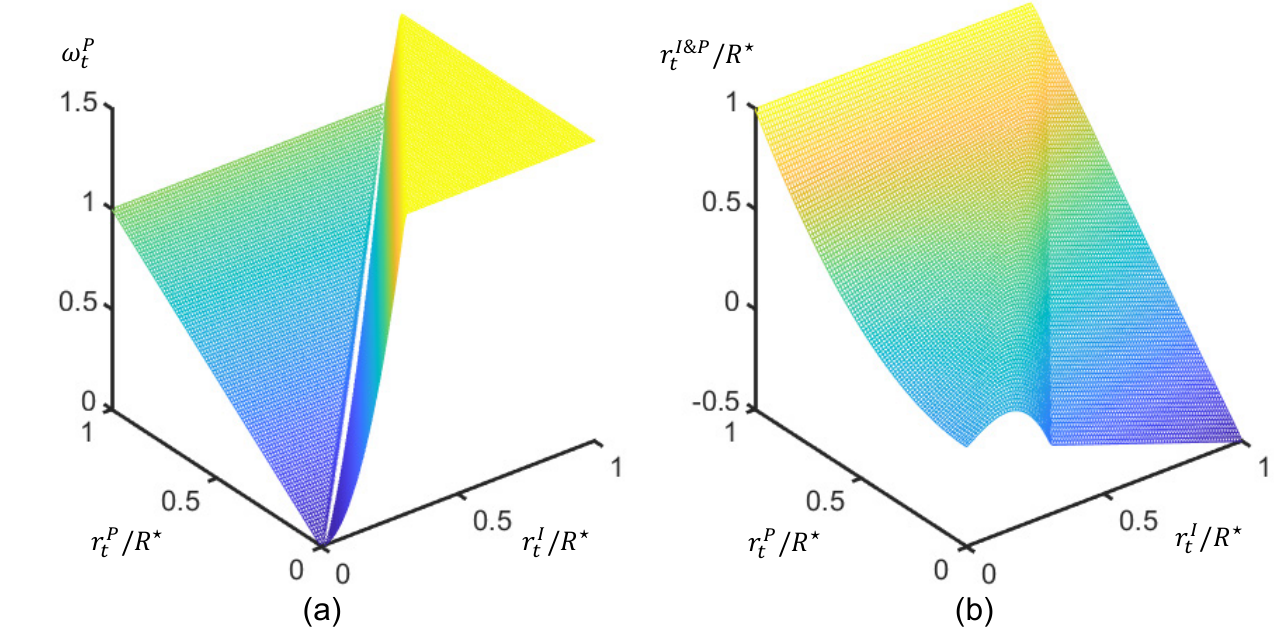}

      \caption{Visualization of (a) $\omega_t^P$ and (b) $r_t^{I\& P}$ according to Eq. (\ref{eq:omega def}). Pursuit of performance is always encouraged, while pursuit of imitation depends on performance.}
      \label{fig:vis_wr}

  \vspace{-1em}
\end{figure}

The key of AdaMimic is the definition of $\omega^P$ function in Eq. (\ref{eq:ada omega}). We define it as
\begin{equation}
    \label{eq:omega def}
    \omega^P_t =\left\{\begin{array}{ll}
\frac{r_t^P}{R^\star} &{\rm\ if\ }r_t^I\le r_t^P, \\
\min \left((4\frac{r_t^I}{R^\star}-3\frac{r_t^P}{R^\star})^2,1.5 \right)&{\rm\ if\ }r_t^I > r_t^P,
\end{array}\right.
\end{equation}
with visualization of $\omega^P_t$ and $r_t^{I\&P}=(1-\omega_t^P)r_t^I+\omega_t^P r_t^P$ in Figure \ref{fig:vis_wr}. A high performance reward is always encouraged, while the critical point for pursuit of imitation is $\omega_t^P=1$. 

Regarding imitation, there are different cases. If both $r_t^P$ and $r_t^I$ are low, imitation is encouraged, because we assume that imitation can help improve the performance in such case. If $r_t^P$ is low while $r_t^I$ is high, imitation is discouraged, with the assumption that excessive imitation is detrimental to performance. If $r_t^P$ is high, imitation is hardly cared about, but excessive imitation is still discouraged if it cannot bring better performance.

\section{Parameterized Bipedal Walking}
\label{sec:methodology}
In this section, we describe our robot platform in \ref{subsec:robot}, and the generation of low-quality gait references in \ref{subsec:gait ref}. The DRL framework of parameterized bipedal walking is presented in \ref{subsec:drl bg}.

\subsection{Robot Model}
\label{subsec:robot}
\begin{figure}[t]
  \centering
  \includegraphics[width=154mm]{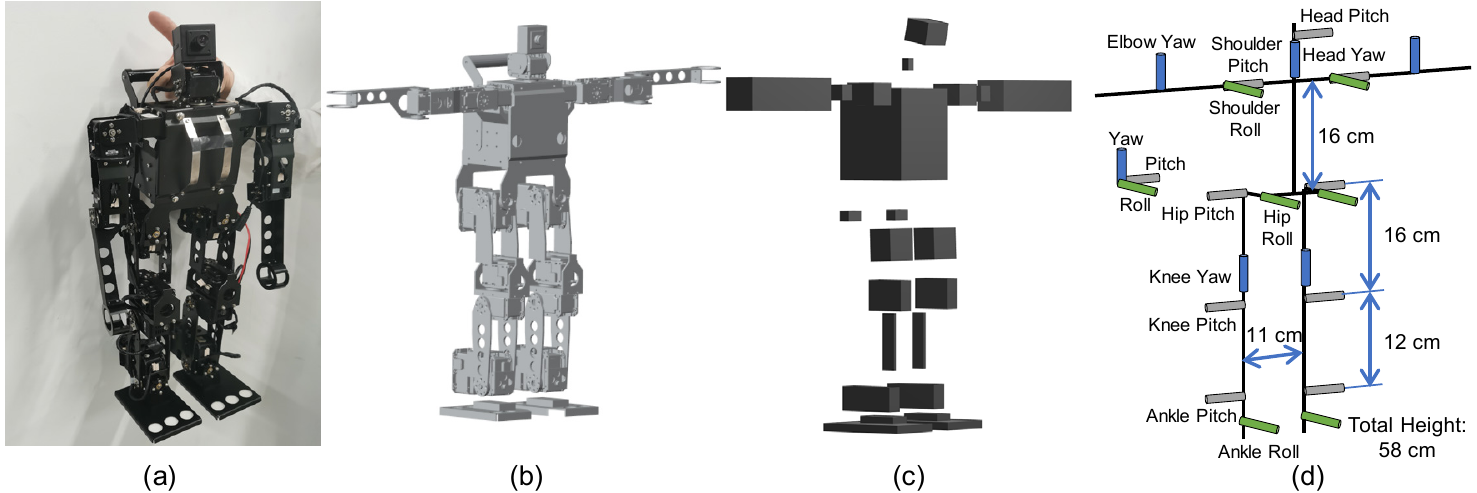}
  \caption{The MOS humanoid platform: (a) the real robot, (b) the mesh in simulation, (c) the simplified collision model, and (d) the distribution of 20 DoFs.}
  \label{fig:showrobot}
  \vspace{-0em}
\end{figure}

Our experiments are conducted on MOS, a self-designed humanoid robot platform. The MOS is 58 cm in height and 5 kg in weight, as is shown in Fig \ref{fig:showrobot}. It has 20 DoFs, including 2 for head, 6 for each leg and 3 for each arm. DYNAMIXEL MX series servos are used for joint actuators. The hardware specifications are shown in Table \ref{tab:joints}. 

\begin{table}[t]
\centering

\resizebox{0.8\textwidth}{!}{%
\begin{tabular}{ccccc}
\hline
\textbf{Part}          & \textbf{DoF}   & \textbf{Type} & \textbf{Stall Torque (N$\cdot$m)} & \textbf{No Load Speed (rev/min)} \\ \hline
                       & Head Pitch  & MX28   & 2.5    & 55   \\
\multirow{-2}{*}{Head} & Head Yaw   & MX28   & 2.5  & 55 \\ \hline
                       & Shoulder Pitch & MX28   & 2.5  & 55   \\
                       & Shoulder Roll  & MX28  & 2.5  & 55    \\
\multirow{-3}{*}{Arm}  & Elbow Yaw     & MX28   & 2.5    & 55     \\ \hline
                       & Hip Roll       & MX106  & 8.4    & 45  \\
                       & Hip Pitch      & MX64   & 6.0   & 63      \\
                       & Knee Yaw     & MX64 & 6.0  & 63   \\
                       & Knee Pitch     & MX64 & 6.0  & 63  \\
                       & Ankle Pitch     & MX64    & 6.0   & 63   \\
\multirow{-6}{*}{Leg}  & Ankle Roll    & MX106   & 8.4    & 45  \\ \hline
\end{tabular}%
}
\caption{Hardware Specifications}
\vspace{-1em}
\label{tab:joints}
\end{table}

\subsection{Gait References}
\label{subsec:gait ref}

\begin{figure}[t]
  \centering
    \includegraphics[width=120mm]{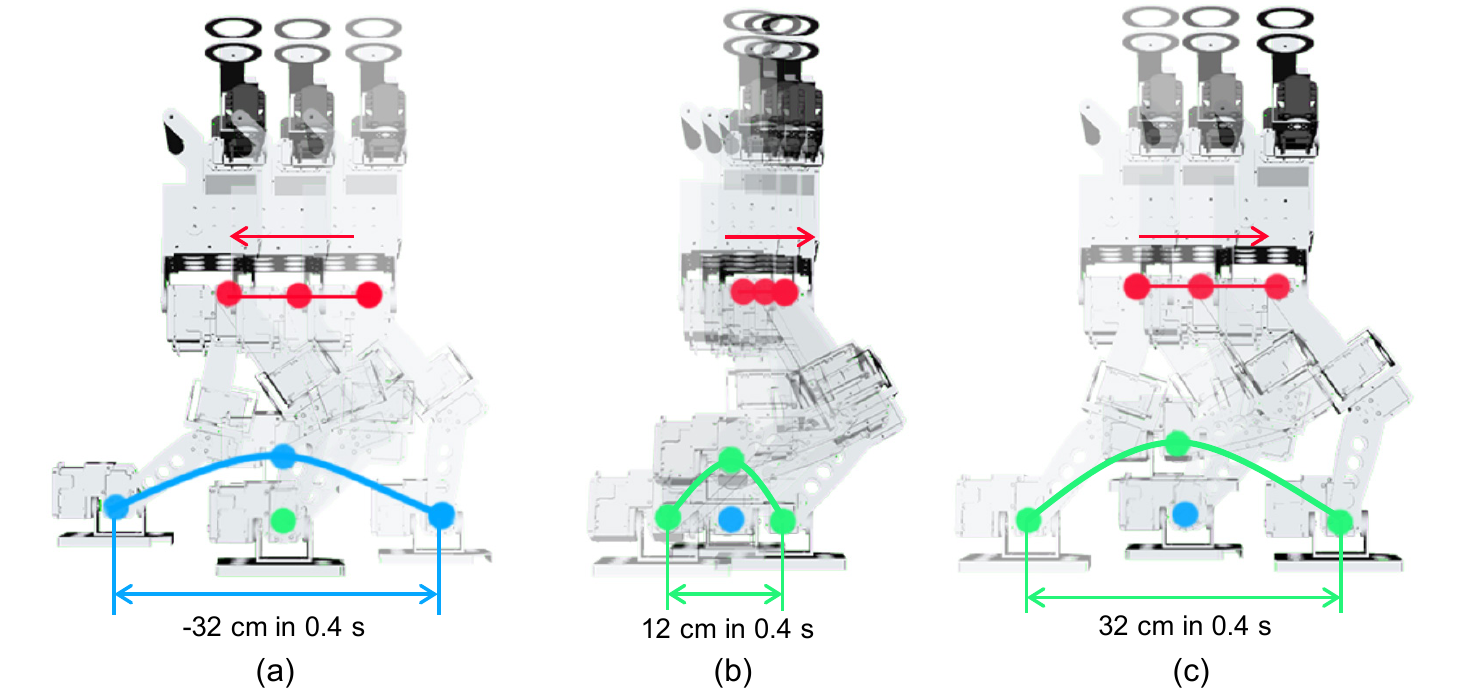}
    \caption{
       Generated gait references for (a) $v_x=-0.4{\rm\ m/s}$, (b) $v_x=0.15{\rm\ m/s}$, and (c) $v_x=0.4{\rm\ m/s}$. The two feet follow symmetric trajectories, thus only one step is shown, and the first step should be a half step. Red points and lines are for the body, blue points and lines are for the left foot, and green points and lines are for the right foot. These references can make the robot fall within 3 steps if directly deployed.
    } 
    \label{fig:ref}
  \vspace{-1em}
\end{figure}

For references, we casually assigned the critical frames and generated the open-loop references by fifth-order Bezier spline interpolation, with joint positions solved by inverse kinematics. We generated gait references of the following parameters for experiments: 1)$v_x=-0.4{\rm\ m/s}, v_y = 0 {\rm\ m/s}, \omega_z = 0{\rm\ rad/s}$, 2) $v_x=0.15{\rm\ m/s}, v_y = 0 {\rm\ m/s}, \omega_z = 0{\rm\ rad/s}$, 3) $v_x=0.4{\rm\ m/s}, v_y = 0 {\rm\ m/s}, \omega_z = 0{\rm\ rad/s}$,
with x-axis pointing forward, y-axis pointing leftward, and z-axis pointing upward. Figure \ref{fig:ref} shows how the references are designed, and the body is assumed to be at a fixed height. These references are all of low quality, i.e., they can not be directly applied to the robot, but are easy to obtain without laborious tuning.

\subsection{Learning of Bipedal Walking}
\label{subsec:drl bg}
For learning, we adopt the massively parallel DRL framework in (\cite{rudin2021learning}). The Proximal Policy Optimization (PPO) algorithm (\cite{schulman2017proximal}) is applied together with generalized advantage estimation (GAE) (\cite {Schulmanetal_ICLR2016}). We adopt the positional control according to (\cite{peng2017learning}), and the policy outputs target joint positions at 50 Hz. The PD controllers generate torques based on the target joint positions and the measured joint positions at 200 Hz. The simulation environment is Isaac Gym (\cite{makoviychuk2021isaac}).

\section{Experiments}
\label{sec:exp}
We experimentally validated our method on the MOS robot introduced in \ref{subsec:robot}, with the DRL framework in \ref{subsec:drl bg}. In this section, the tasks and settings of our experiments are respectively presented in \ref{subsec:tasks} and \ref{subsec:settings}, and the results are displayed in \ref{subsec:results}.
\subsection{Tasks}
\label{subsec:tasks}
As is described in \ref{subsec:gait ref}, we generated three low-quality references, respectively with $v_x=-0.4{\rm\ m/s}$, $v_x=0.15{\rm\ m/s}$, and $v_x=0.4{\rm\ m/s}$. All of them have $v_y = 0 {\rm\ m/s}, \omega_z = 0{\rm\ rad/s}$.

To show the applicability of our method, we designed different tasks below:
\newline 1) \textbf{From the same reference to different targets}
\begin{enumerate}[label=1.\arabic*.,leftmargin=2\parindent]
    \item learning to walk at $v_x=0.4{\rm\ m/s}, v_y = 0 {\rm\ m/s}, \omega_z = 0{\rm\ rad/s}$ from reference $v_x=0.4{\rm\ m/s}$;
    \item learning to walk at $v_x=0.6{\rm\ m/s}, v_y = 0 {\rm\ m/s}, \omega_z = 0{\rm\ rad/s}$ from reference $v_x=0.4{\rm\ m/s}$;
    \item learning to walk at $v_x=0.5{\rm\ m/s}, v_y = 0.2 {\rm\ m/s}, \omega_z = -1{\rm\ rad/s}$ from reference $v_x=0.4{\rm\ m/s}$.
\end{enumerate}
2) \textbf{From different references to the same target}
\begin{enumerate}[label=2.\arabic*.,leftmargin=2\parindent]
    \item learning to walk at $v_x=0.6{\rm\ m/s}, v_y = 0 {\rm\ m/s}, \omega_z = 0{\rm\ rad/s}$ from reference $v_x=0.4{\rm\ m/s}$;
    \item learning to walk at $v_x=0.6{\rm\ m/s}, v_y = 0 {\rm\ m/s}, \omega_z = 0{\rm\ rad/s}$ from reference $v_x=0.15{\rm\ m/s}$;
    \item learning to walk at $v_x=0.6{\rm\ m/s}, v_y = 0 {\rm\ m/s}, \omega_z = 0{\rm\ rad/s}$ from reference $v_x=-0.4{\rm\ m/s}$.
\end{enumerate}
Note that task 2.1 is the same as task 1.2, so there are 5 tasks and 3 targets in total. It is also worth mentioning that a speed of $0.6{\rm \ m/s}$ is close to the limit of the MOS robot which is $0.58{\rm\ m}$ tall and can hardly jump.

\subsection{Settings}
\label{subsec:settings}

For learning, we adopted the framework in (\cite{rudin2021learning}), as is mentioned in \ref{subsec:drl bg}, and kept most of the configurations as original. Changed configurations are: $n_{\rm robots}=4096$, $n_{\rm steps}=96$, plain terrains without measurements sampled from the grid.

\begin{table}[t]
\begin{minipage}[b]{.35\linewidth}
    \centering
    \resizebox{\textwidth}{!}{
    \begin{tabular}{|ll|}
            \hline
            \multicolumn{2}{|c|}{Symbols} \\ \hline
            Body height & $h$ \\
            Joint positions & $q$ \\
            Joint velocities & $\dot{q}$ \\
            Joint accelerations & $\ddot{q}$ \\
            Joint torques & $\tau$\\
            Target joint positions (action) & $a$ \\
            Difference of the action & $\delta a$\\
            Reference joint positions &$q^r$\\
            Base linear velocity & $v$\\
            Base angular velocity & $\omega$ \\
            Commanded base linear velocity & $v^\star$\\
            Commanded base angular velocity & $\omega^\star$ \\
            Environment time step & $dt$\\
            \hline
        \end{tabular}
        }
\end{minipage}
\begin{minipage}[b]{.65\linewidth}
    \centering
    \resizebox{\textwidth}{!}{
        \begin{tabular}{|rl|}
            \hline
            \multicolumn{2}{|c|}{Imitation reward $r_t^I$, $R^\star=1dt$, sum of below} \\ \hline
            Imitation reward for joint positions & $1dt\cdot{\rm RBF}(q,q^r,1.5)$ \\ 
            \hline \hline
            \multicolumn{2}{|c|}{Performance reward $r_t^P$, $R^\star=1dt$, sum of below} \\ \hline
            Linear velocity tracking & $0.75dt\cdot {\rm RBF}([v_x,v_y],[v_x^\star,v_y^\star],0.16)$ \\
            Angular velocity tracking & $0.25dt\cdot {\rm RBF}(\omega_z,\omega_z^\star,2.5)$ \\
            \hline \hline
            \multicolumn{2}{|c|}{Regularization reward $r_t^R$, sum of below} \\ \hline
            Torque regularization & $-1\times 10^{-6}dt\cdot \|\tau\|_{2}^{2}$ \\ 
            Joint acceleration regularization & $-2\times 10^{-8}dt\cdot \|\ddot{q}\|_{2}^{2}$ \\ 
            Z-axis Linear velocity regularization & $-2\times 10^{-2}dt\cdot v_z^{2}$ \\ 
            Roll rate and pitch rate regularization & $-2\times 10^{-2}dt\cdot (\omega_x^2+\omega_y^2)$ \\ 
            Action rate regularization & $-1\times 10^{-4}dt\cdot \|\delta a\|_{2}^{2}$ \\ 
            \hline \hline
            \multicolumn{2}{|c|}{Other reward $r_t^O$, sum of below} \\ \hline
            Alive reward & -50 if $h<0.33{\rm \ m}$ then reset, else 0 \\ 
            \hline
        \end{tabular}}
\end{minipage}

\caption{Reward Terms}
\label{tab:reward}
\vspace{-1em}
\end{table}

The reward terms are displayed in Table \ref{tab:reward}. The RBF function applied in the table is
\begin{equation}
    \label{eq:rbf}
    \operatorname{RBF}(x, y, \sigma^2)=\exp \left(-\frac{\|x-y\|_{2}^{2}}{\sigma^2}\right).
\end{equation}
And the reward function is the sum of the weighted terms according to Eq. (\ref{eq:dpm form}), with $\omega^R=\omega^O=1$. 

For comparison, we tried 5 strategies for the other two weights:
\begin{enumerate}
    \item $\omega_t^P$ is determined by Eq. (\ref{eq:omega def}) and $\omega_t^I=1-\omega_t^P$, following our AdaMimic formulation, abbreviated to "Ada";
    \item Fixed $\omega_t^P = 0.2$ and $\omega_t^I=1-\omega_t^P=0.8$, following the DeepMimic formulation, abbreviated to "DM0.2";
    \item Fixed $\omega_t^P = 0.5$ and $\omega_t^I=1-\omega_t^P=0.5$, abbreviated to "DM0.5";
    \item Fixed $\omega_t^P = 0.8$ and $\omega_t^I=1-\omega_t^P=0.2$, abbreviated to "DM0.8";
    \item Fixed $\omega_t^P = 1$ and $\omega_t^I=1-\omega_t^P=0$, i.e, no imitation, abbreviated to "NI".
\end{enumerate}
For "NI", different tasks are only determined by different targets, so there are only 3 tasks (task 1.1, 1.2, and 1.3) for "NI".

\subsection{Results}
\label{subsec:results}

\begin{figure}[t]
  \centering
  \includegraphics[width=150mm]{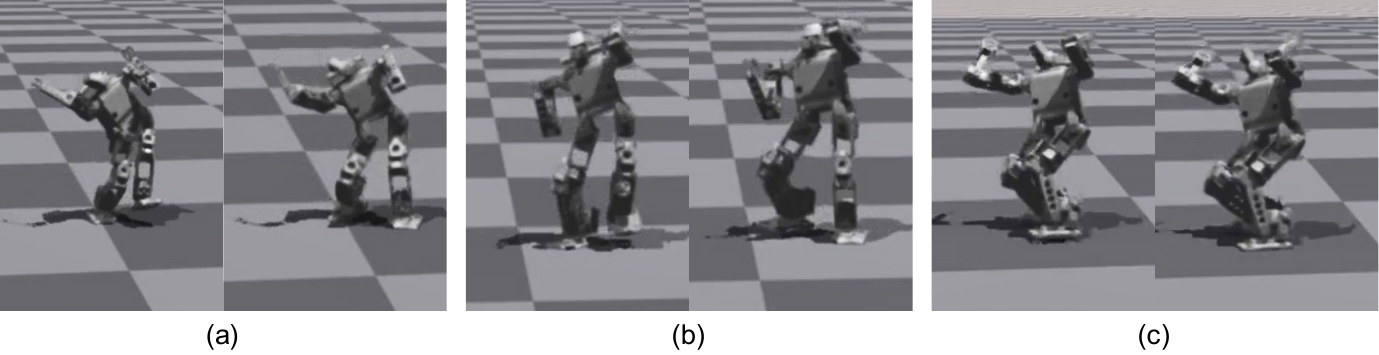}
  \caption{Snapshots of the learned gait patterns without imitation after 1000 iterations. (a) for task 1.1, (b) for task 1.2, and (c) for task 1.3.}
  \label{fig:ni_display}
\bigskip
\vspace{-0em}
  \centering
  \includegraphics[width=150mm]{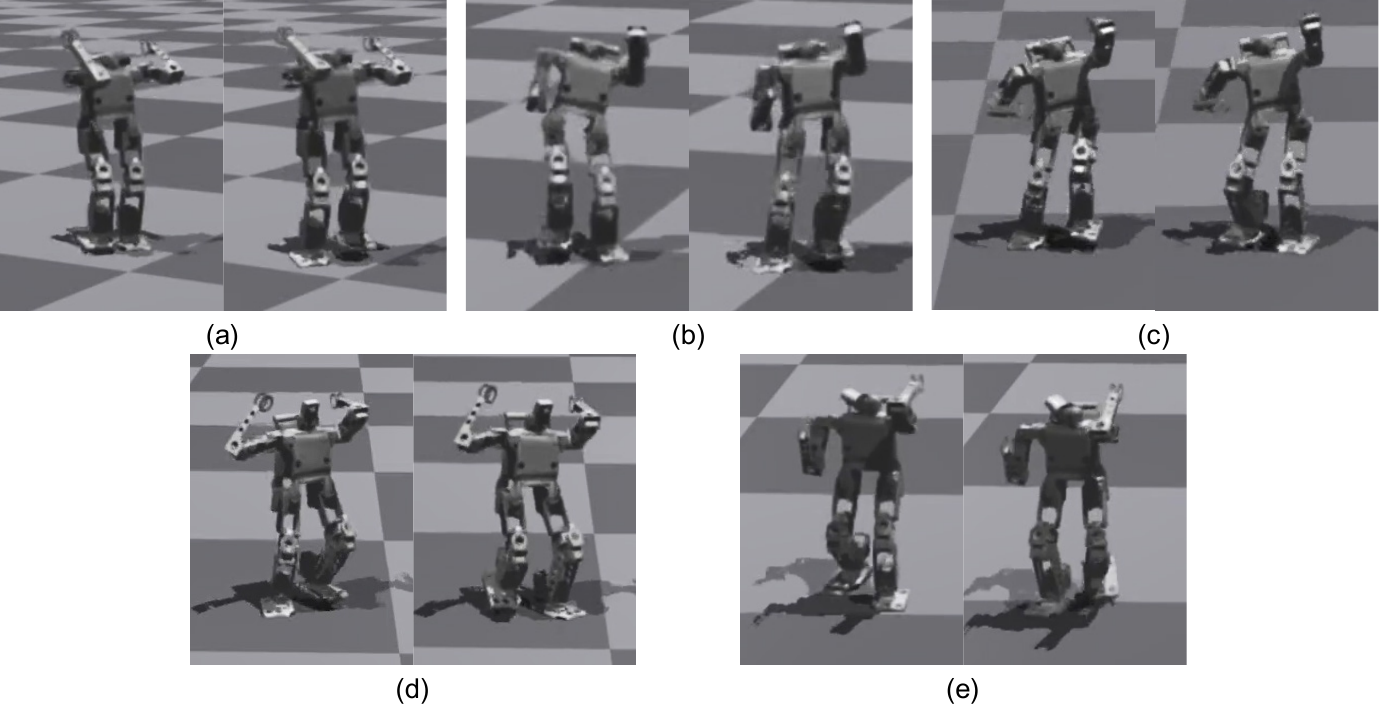}
  \caption{Snapshots of the learned gait patterns through AdaMimic after 1000 iterations. Upper body joints are not included in imitation. (a) for task 1.1, (b) for task 1.2, (c) for task 1.3, (d) for task 2.2, and (e) for task 2.3. The patterns are more reasonable and elegant compared to those in Figure \ref{fig:ni_display}. }
  \label{fig:ada_display}
\end{figure}

The essential distinction between "NI" and other strategies is that the robot could not learn reasonable gait patterns. Therefore, for "NI", we only show some snapshots in Figure \ref{fig:ni_display}. For comparison, snapshots of gait patterns learned through "Ada" are also displayed in Figure \ref{fig:ada_display}.

\begin{figure}[t]
\centering

  \includegraphics[width=150mm]{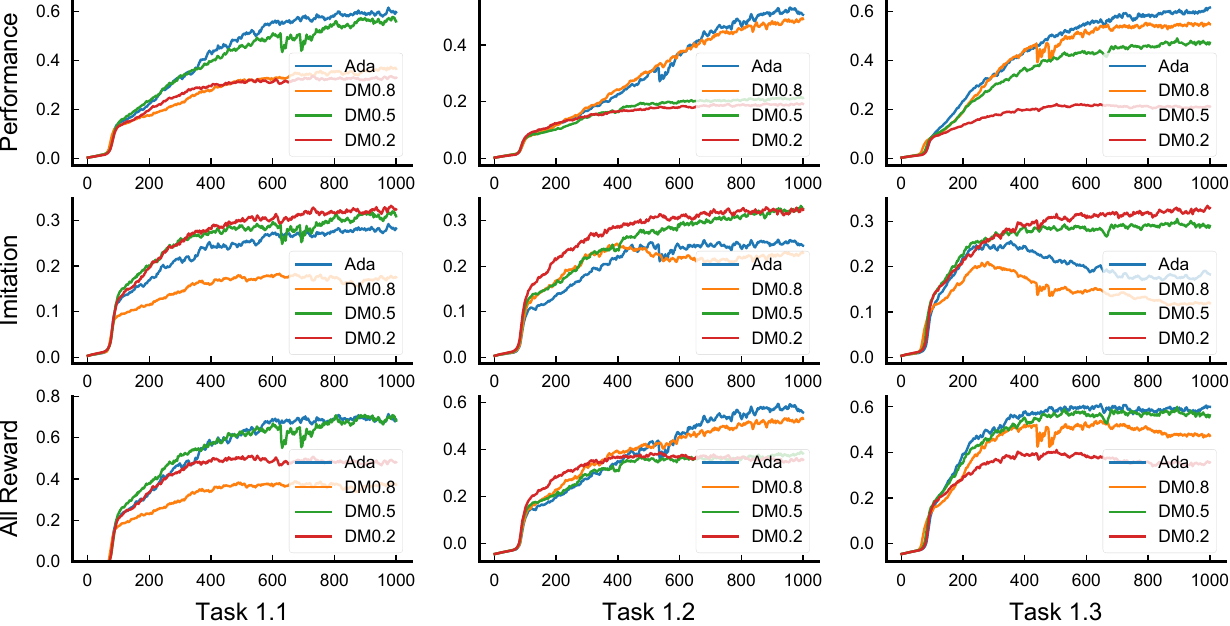}
  \caption{Learning curves for task 1.1, 1.2, and 1.3. Empirically, a "Performance" larger than 0.5 means that the robot is going to achieve the target.}
  \label{fig:curves1}

\bigskip
\vspace{-0em}

  \centering
  \includegraphics[width=150mm]{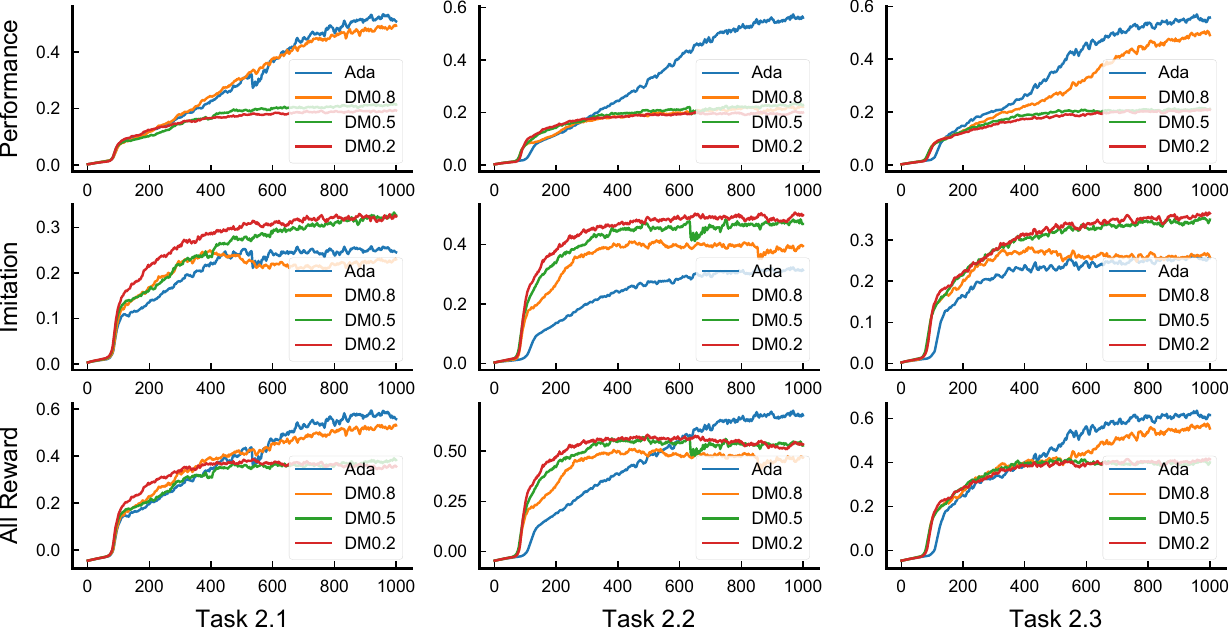}
  \caption{Learning curves for task 2.1, 2.2, and 2.3. Empirically, a "Performance" larger than 0.5 means that the robot is going to achieve the target.}
  \label{fig:curves2}
\vspace{-1em}
\end{figure}

Learning curves of other strategies are shown in Figure \ref{fig:curves1} and Figure \ref{fig:curves2}. X-axis and Y-axis of the plots respectively correspond to the "iterations" and the "Episode Reward" in the framework in (\cite{rudin2021learning}). The "Performance" reward demonstrates how well the target is achieved, and our proposed AdaMimic successfully outperformed the DeepMimic formulation. For different references and targets, the best DeepMimic weight changes, but the AdaMimic strategy does not require the manual adjustment of weights.

\begin{figure}[t]
\centering
  \includegraphics[width=150mm]{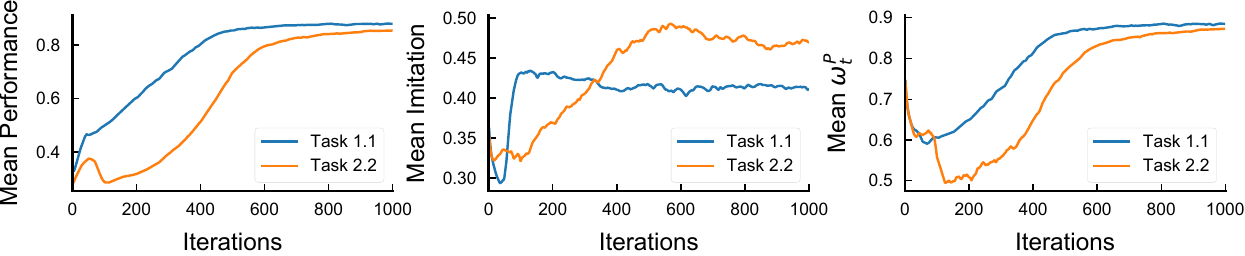}
  \caption{Changes of (a) mean performance reward, (b) mean imitation reward, and (c) mean $\omega_t^P$ for task 1.1 and tasks 2.2.}
  \label{fig:curves3}
\vspace{-1em}
\end{figure}

For different tasks, the switch time between pursuit of imitation and performance can be different, but the pattern is similar, as is shown in Figure \ref{fig:curves3}. The imitation is firstly encouraged with a low $\omega_t^P$, but as the performance reward increases, the imitation reward slowly decreases. The "mean" terms in Figure \ref{fig:curves3} are averaged over steps, instead of over the horizon as in Figure \ref{fig:curves1} and Figure \ref{fig:curves2}.

\section{Conclusion and Future Work}
\label{sec:conclusion}
In this paper, we proposed the AdaMimic formulation for imitation learning of robot behaviors from low-quality references. Based on infeasible references, policies of parameterized bipedal walking were learned through AdaMimic for our MOS humanoid robot. We claim that the references were generated without manual tuning, and the weights of imitation and performance can be automatically adjusted during the learning process, which finally serves to pursue better performance.

To obtain ideal references for various locomotion tasks, laborious motion capturing or manual tuning of parameters is usually unavoidable. AdaMimic should change this situation because it only requires simple references without tuning.

In the future, we will further use simple references to help our MOS robot learn omnidirectional bipedal locomotion and achieve the ability to travel across difficult terrains. The learned policy will be tested on the real robot. Extensions to other locomotion tasks such as fall recovery and soccer dribbling will also be explored.



\bibliography{ref.bib}

\end{document}